\documentclass[a4paper]{esannV2}

\usepackage[dvips]{graphicx}
\usepackage[latin1]{inputenc}
\usepackage{amssymb,amsmath,array}

\usepackage{subcaption,siunitx,booktabs}

\usepackage{hyperref}
\usepackage{multirow}
\usepackage{booktabs}
\usepackage{scrextend}
\usepackage{tablefootnote}
\usepackage{epstopdf}

\begin{document}
\title{End-to-end Keyword Spotting using Xception-1d}


\author{Iván Vallés-Pérez, Juan Gómez-Sanchis, Marcelino Martínez-Sober \\ Joan Vila-Francés, Antonio J. Serrano-López, Emilio Soria-Olivas
%

\vspace{.3cm}\\
%
Intelligent Data Analysis Laboratory (IDAL), University of Valencia,\\ Avenida de la Universitat s/n 46100 Burjassot, Valencia, Spain
}

\maketitle

\begin{abstract}
The field of conversational agents is growing fast and there is an increasing need for algorithms that enhance natural interaction. In this work we show how we achieved state of the art results in the Keyword Spotting field by adapting and tweaking the \textit{Xception} algorithm, which achieved outstanding results in several computer vision tasks. We obtained about 96\% accuracy when classifying audio clips belonging to 35 different categories, beating human annotation at the most complex tasks proposed.

\end{abstract}

\vspace{-6pt}
\section{Introduction}
\vspace{-5pt}

Virtual assistants are usually implemented into small devices with low power specifications. Generally, a low latency is required to avoid harming users' experience. For that reason, choosing a lightweight solution is critical, as this is in essence a \textit{TinyML} problem \cite{sanchez2020}. Using the cloud for processing audio commands can be pricey and increase latency. Therefore, at least the models that recognize the most common words in a limited vocabulary, such as \textit{KeyWord Spotting} (KWS), should be implemented locally. This work focuses on Deep Learning (DL) models to increase the accuracy of KWS models.

The current virtual assistants are still not as accurate as humans at identifying voice commands \cite{Michaely2017}. Although many efforts have been made, there is still room for improvement. In particular, bidirectional recurrent models with attention have been used \cite{Andrade2018}. Gated convolutional Long-Short Term Memory (LSTM) structures have also proven useful by other authors \cite{Wang2018}.  Due to their architecture, Convolutional Neural Networks (CNNs) provide a good approach to optimize computational resources for KWS. In \cite{Tara2015}, CNNs outperform other deep neural networks (DNNs) architectures, at the KWS task. Other works focus on the hardware implementation of neural networks for KWS \cite{Zhang2017}, comparing their accuracy, memory usage and computation efficiency. Transfer learning has also been tested in this domain \cite{McMahan2018} showing substantial improvements in accuracy.

This work presents \textit{Xception-1d}, a CNN architecture based on \textit{Xception} \cite{FChollet2017}, to tackle the speech commands recognition problem.  Our contribution is summarized as follows: (1) design of a \textit{depthwise separable} CNN-based architecture with better results than the existing benchmarks (including human annotators), (2) description of an efficient methodology for augmenting audio data (multiplying its size by 5) (3) human performance quantification to use it as an additional baseline, (4) creation of a public repository to foster reproducibility.

The article is structured as follows. Section \ref{sec:MM} introduces the data and algorithms used. Section \ref{sec:results} shows the results achieved by the proposed algorithm. Section \ref{sec:discussion} provides a discussion of the results, and section \ref{sec:conclusion} summarizes the main conclusions.

\section{Data set} \label{sec:MM}

We used the \textit{Google Tensorflow speech commands data set} \cite{Warden2018} along this study. It consists of a collection of 1-second long annotated utterances containing short words, recorded by thousands of non-professional speakers. The subjects were asked to say a chain of words during a minute. Then, the recordings were split in clips of one second by extracting the loudest sections of the signal \cite{Warden2018}.

Two different versions of the data set are available (referred as V1 and V2 hereafter), containing 64,721 and 105,829 audio clips of one second (each one containing the recording of a single voice command), respectively, with a sample rate of 16 kHz and 16-bit resolution and stored in \textit{wav} format. The first data version has up to 30 different voice commands while the second one has 35 of them. The classes proportions are fairly balanced, although we are not including the detail figure due to space constraints.

The data sets contain recordings of the following words: ``left'',
``right'',
``yes'',
``no'',
``down'',
``up'',
``go'',
``stop'',
``on'',
``off'',
``zero'',
``one'',
``two'',
``three'',
``four'',
``five'',
``six'',
``seven'',
``eight'',
``nine'',
``dog'',
``cat'',
``wow'',
``house'',
``bird'',
``happy'',
``sheila'',
``marvin'',
``bed'',
``tree'',
``visual'',
``follow'',
``learn'',
``forward'',
``backward''. The last 5 words being exclusive of V2 of the data set

Four different tasks have been defined in order to benchmark the proposed algorithm: (1) \textit{35-words-recognition}: consists of classifying all the words. (2) \textit{20-commands-recognition}: classifying a subset of 20 words (robot commands + numbers), grouping the rest of the utterances in an ``unknown'' class. (3) \textit{10-commands-recognition}: categorizing a subset of 10 words (robot commands), grouping the other words into the ``unknown'' class. (4) \textit{left-right-recognition}: categorizing the ``left'' and ``right'' commands, while the rest of clips are grouped under the ``unknown'' category.

\subsection{Data augmentation}
Five different distorted versions of each clip have been generated. These distortions consist of a set of transformations being applied together over all the clips randomizing their parameters and intensities. The different distortions used in this step are: (1) resampling: extending/contracting the clip length, affecting its pitch (2) saturation: application of a random amplification, (3) time offset: displacement in time of the audio clip, (4) white noise addition: addition of a random amplitude Gaussian noise, (5) Pitch shift: application of pitch distortion \cite{Szymon2016}.

All the distortions are applied together with random intensities only to the training data, producing 5 new transformations of the original recordings with high variability of results. These new versions are appended to the original data set.

\section{Methods}
We propose the use of a variant of the \textit{Xception architecture}  described by \textit{François Chollet} in 2017 \cite{FChollet2017}, due to its low computational cost. This model recently achieved state of the art results in multiple computer vision tasks \cite{Liu2019, Song2018}.

\subsection{\textit{Xception-1d} architecture}
\begin{figure}[ht]
	\centering
	\includegraphics[width=0.7\linewidth]{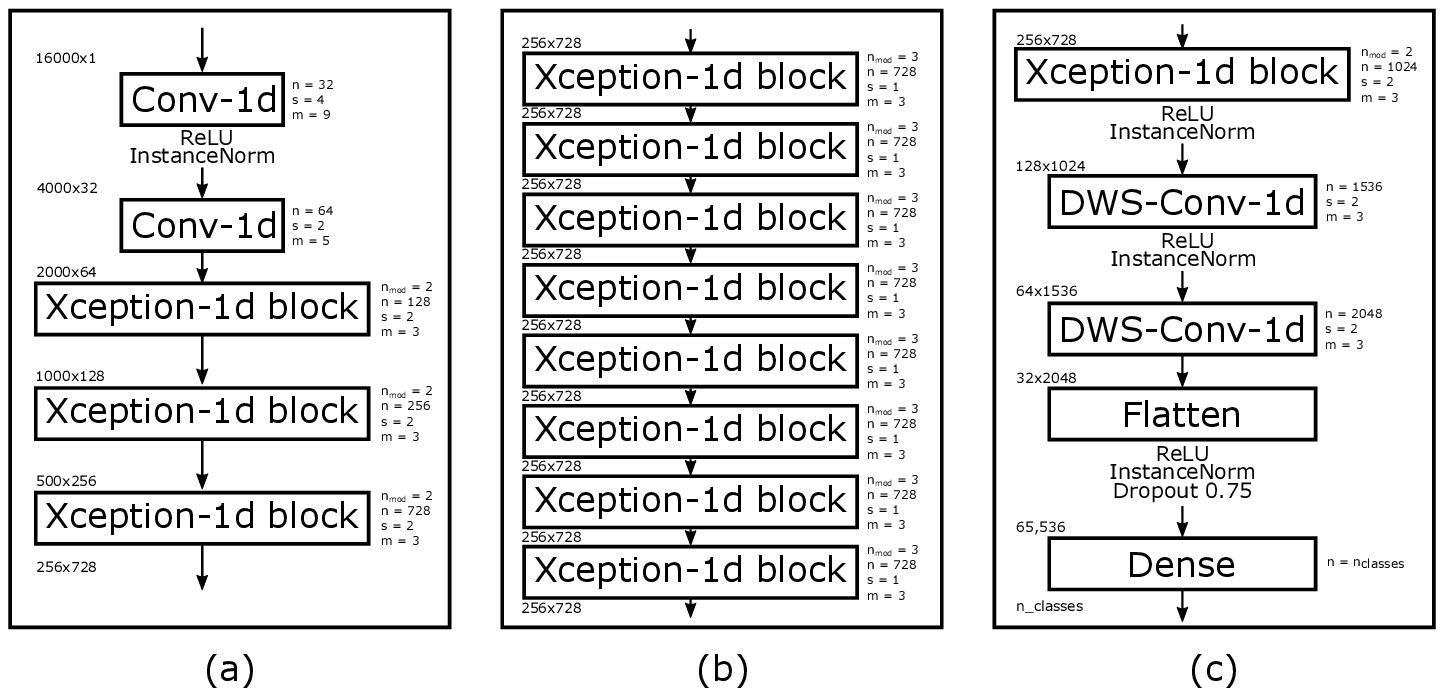}
	\includegraphics[width=0.29\linewidth]{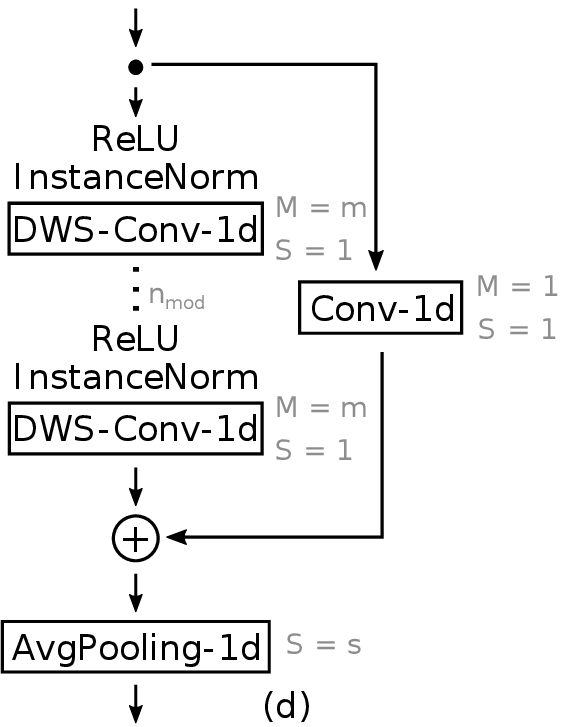}
	\caption{\textit{Xception-1d} architecture where $n$ is the number of channels, $s$ is the stride of the convolutions, $m$ is the size of the convolution filters, $n_{mod}$ is the number of \textit{depthwise convolutions} and $n_{classes}$ is the number of outputs of the network. The architecture has 3 main modules: (a) the entry module is responsible for adapting the input wave into a condensed representation, (b) the middle module is responsible for learning the abstract representation  and (c) the classification module is responsible for mapping the extracted features into the number of outputs required for every task. The Xception-1d block is depicted in (d).}
	\label{fig:arch}
\end{figure}

The \textit{Xception-1d} architecture (see Figure \ref{fig:arch}) contains the next modifications over the original architecture \cite{FChollet2017}: (1) we turned 2D convolutions into 1D, (2) we used Instance Normalization to normalize the outputs of each convolution \cite{Zheng2018}. The input of the model is an audio clip in temporal domain.

The proposed architecture exploits the gain in efficiency of the \textit{depthwise separable convolution} operation \cite{Yunhui2019} over the \textit{regular convolution} in the same way original \textit{Xception} does \cite{FChollet2017}. The number of operations required by this layer is $\frac{1}{N} \cdot \frac{1}{S}$ times the amount in a regular convolution \cite{Howard2017}, where $S$ is the size of the \textit{depthwise convolution} filters and $N$ is the number of output channels after the \textit{pointwise convolution} is applied.

Considering that the last dense layer can be prone to overfitting due to the high number of parameters ($\sim 65,000$), a strong dropout \cite{Goodfellow2016} has been applied after the last convolution ($p = 75\%$). In addition, a small \textit{weight decay} \cite{Goodfellow2016} has been applied over all the network weights ($\lambda = 10^{-3}$) in order to enhance regularization. \textit{Adam} optimizer has been used to train the network \cite{Kingma14}. The initial learning rate ($\eta = 10^{-4}$), has been decreased by a factor of $\frac{1}{2}$ when no improvement was observed, with a patience of 4 epochs.

\section{Results} \label{sec:results}
\begin{figure}[ht]
	\centering
	\includegraphics[width=0.7\linewidth]{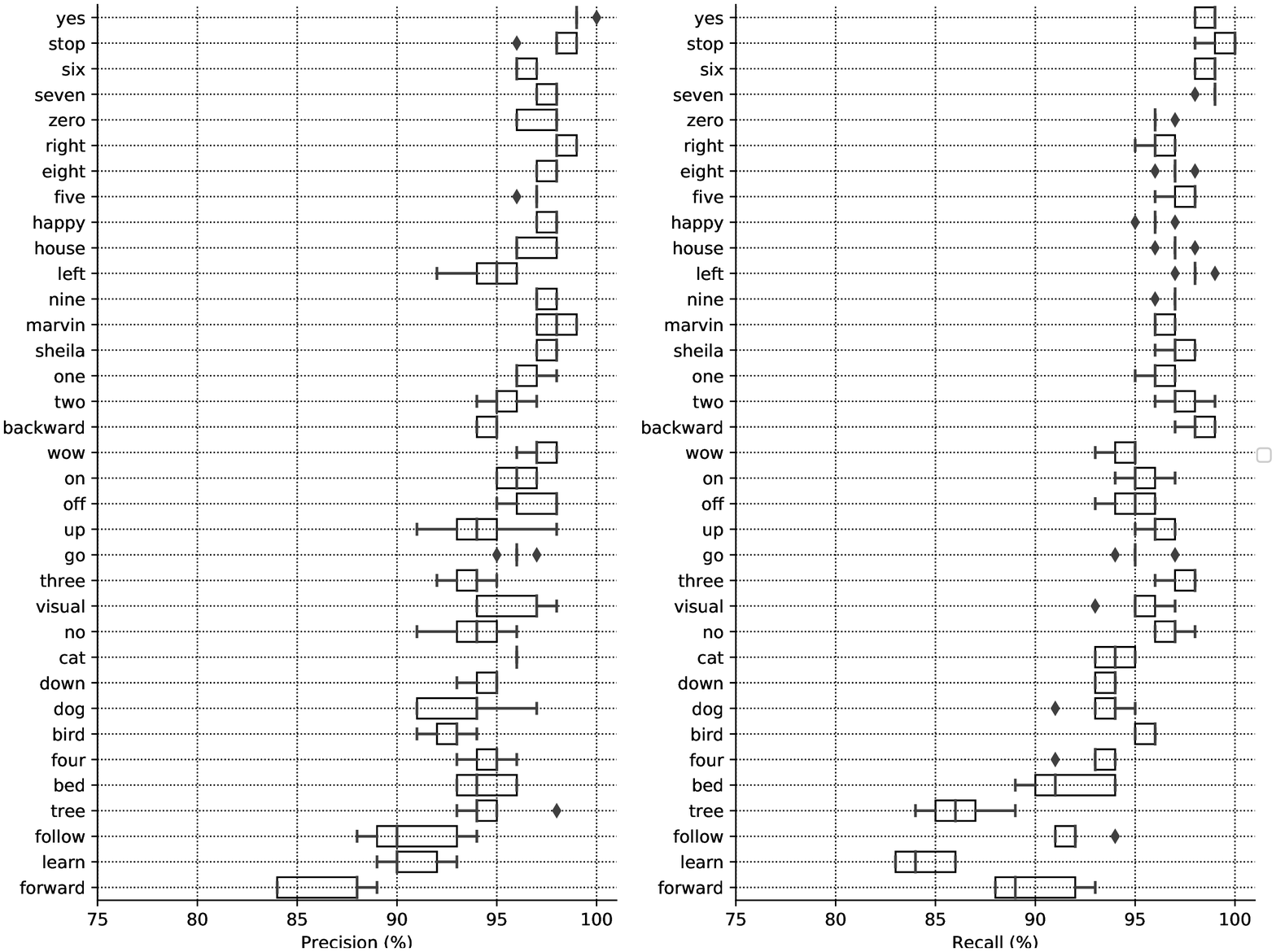}
	\caption{Precision and recall for each class using the \textit{35-words-recognition} model trained with data V2. Classes are sorted by descending f1-score.}
	\label{fig:boxplot35002}
\end{figure}

\vspace{-5pt}
The train/development/test split provided by the authors of the data set \cite{Warden2018} has been adopted as \textit{cross validation} (CV) setting in order to facilitate future benchmarking efforts. We hold 16,000 and 9,981 samples for development and 16,000 and 11,005 samples testing purposes, in V1 and V2 respectively. Speakers in the training set are not present in the test set. The model has been trained for 50 epochs in each case, with a batch size of 32 clips, and the weights of the epoch that achieved the best performance in the development set were used to report the performance of the algorithm. The results shown in this section have been measured over the test set. Five different models have been trained for each task in order to explore and report the effect of different random weights initializations. With the aim of providing a baseline, human performance has been measured by 4 human subjects, each labeling $\sim 1000$ commands. These results are reported in Table \ref{tab:comparative} along with the performance of the proposed algorithm and benchmarks. The source code is available here: \url{https://github.com/ivallesp/Xception1d}

Besides the global results, Figure \ref{fig:boxplot35002} shows the precision and the recall obtained for the most complex model (\textit{35-words-recognition} for data version V2). 
\vspace{-20pt}

\begin{table}[ht]
	\centering
	\caption{Accuracy (mean $\pm$ standard deviation) for our proposal, benchmarks and human annotation. Best results per task in bold. Results significantly beating human performance ($\alpha = 0.05$) have been tagged with a star mark (*).}
	\label{tab:comparative}
	
	\vspace{-10pt}
	\begin{subtable}{1\textwidth}
		\caption{Results for version 1 of the data set.}
		\vspace{-5pt}
		\centering
		\footnotesize
		\scalebox{1}{
			\begin{tabular}{lcccccc}
				\hline
				& \cite{Andrade2018} & \cite{McMahan2018}\footref{foot:max} & \cite{Warden2018} & \textit{Xception-1d} & Human & p-value\footref{foot:ttest}      \\ \hline
				35-words & 94.30 & 84.35 & - & \textbf{95.85 $\pm$ 0.12 *} & 94.15 $\pm$ 1.03  & $1.46\cdot 10^{-2}$\\
				20-commands & 94.10 & 85.52 & - & \textbf{95.89 $\pm$ 0.06 *} & 94.56 $\pm$ 0.98  & $3.14\cdot 10^{-2}$\\
				10-commands & 95.60 & - & 85.40 & \textbf{97.15 $\pm$ 0.03} & 97.22 $\pm$ 0.85 & $8.75\cdot 10^{-1}$\\
				left-right & \textbf{99.20} & 95.32 & - & 98.96 $\pm$ 0.09 & 99.54 $\pm$ 0.16 & $5.24\cdot 10^{-4}$ \\ \bottomrule
		\end{tabular}}
		
	\end{subtable}
\vspace{-10pt}
\bigskip

	\begin{subtable}{1\textwidth}
		\caption{Results for version 2 of the data set.}
		\vspace{-5pt}
		\centering
		\footnotesize
		\scalebox{1}{
			\begin{tabular}{lcccccc}
				\hline
				& \cite{Andrade2018} & \cite{Zhang2017}\footnote{the  best results obtained among all the trials performed by the autors have been selected \label{foot:max}} & \cite{Warden2018} &           \textit{Xception-1d}           &     Human    & p-value \footnote{Student's t-test for the comparison of two means. $\alpha = 0.05$ \label{foot:ttest}} \\ \hline
				35-words    &               93.90               &               -               &            -             & \textbf{95.85 $\pm$ 0.16 *} & 94.15 $\pm$ 1.03 & $1.50\cdot 10^{-2}$ \\
				20-commands &               94.50               &               -               &            -             & \textbf{95.96 $\pm$ 0.16 *} & 94.56 $\pm$ 0.98 & $2.70\cdot 10^{-2}$\\
				10-commands &               96.90               &             95.40             &          88.20           & \textbf{97.54 $\pm$ 0.08}  & 97.22 $\pm$ 0.85 & $4.84\cdot 10^{-1}$ \\
				left-right  &          \textbf{99.40}           &               -               &            -             & 99.25 $\pm$ 0.07  & 99.54 $\pm$ 0.16 & $1.27\cdot 10^{-2}$\\ \bottomrule\\
		\end{tabular}}
	\end{subtable}
\end{table}

\vspace{-20pt}
\section{Discussion} \label{sec:discussion}
\vspace{-5pt}

\textit{Xception-1d} offers better performance than the existing methods in the literature for three out of the four tested tasks. In the only task where \textit{Xception-1d} did not achieve the best results (left-right), the leading method was the one proposed by Andrade \textit{et al}. \cite{Andrade2018} which was only marginally better ($<$0.5\% performance difference on the test set) than the presented method. \textit{Xception-1d} even surpassed human performance (with statistical significance level) in the two first tasks, including the most difficult one (\textit{35-words-recognition}).

With regard to per-class accuracy, Figure \ref{fig:boxplot35002} shows that the algorithm performs generally well for all the classes in the most difficult scenario (35-words task) as the majority of precision and recall values lay between 90-100\%. Nonetheless, we found that the algorithm has more difficulties differentiating some groups of similar words like the following pairs: ``three'' and ``tree'', ``follow'' and ``four'', ``bed'' and ``bird'', etc. No comparison with other existing models has been included because such detailed results have not been found in the related work.

Finally, our findings show that the per-class performance of the speech commands in the \textit{left-right} task is lower than for the other tasks (in particular the recall values). However, the per-class performance for these two classes was higher when they were included in a multiclass classification task like the \textit{35-words-recognition} task. This fact suggests that auxiliary tasks (in this case the 35-words vocabulary task) can benefit primary tasks (left-right), as more features may be extracted from more complex tasks. This hypothesis has been proven for Reinforcement Learning \cite{Jaderberg2016}, but may also apply to DL efforts.

\vspace{-8pt}
\section{Conclusion} \label{sec:conclusion}
\vspace{-6pt}
This work shows how a neural network which succeeded in the computer vision field, with an adaption and a set of tweaks, is able to surpass human performance at a speech recognition task with limited vocabulary achieving state of the art results. This is why we suggest \textit{Xception-1d} as the \textit{de facto} architecture when facing a voice command recognition task with restricted vocabulary,  considering the computing power is not a limiting factor due to its small size. The algorithm presented can have multiple  applications for improving voice-controlled systems.

\rule{4cm}{0.4pt}

\textit{This work has been partially funded by the Spanish ministry of science and innovation project PID2019-107347RR-C33}.

\begin{footnotesize}

\bibliographystyle{unsrt}
\bibliography{doc}

\begin{thebibliography}{10}

\bibitem{sanchez2020}
R.~{Sanchez-Iborra} and A.~F. {Skarmeta}.
\newblock Tinyml-enabled frugal smart objects: Challenges and opportunities.
\newblock {\em IEEE Circuits and Systems Magazine}, 20(3):4--18, 2020.

\bibitem{Michaely2017}
Assaf~Hurwitz Michaely, Xuedong Zhang, Gabor Simko, Carolina Parada, and Petar
  Aleksic.
\newblock Keyword spotting for google assistant using contextual speech
  recognition.
\newblock In {\em 2017 IEEE Automatic Speech Recognition and Understanding
  Workshop (ASRU)}, pages 272--278. IEEE, 2017.

\bibitem{Andrade2018}
Douglas Coimbra~de Andrade, Sabato Leo, Martin Loesener Da Silva~Viana, and
  Christoph Bernkopf.
\newblock A neural attention model for speech command recognition.
\newblock {\em Computing Research Repository CoRR}, August 2018.

\bibitem{Wang2018}
Dong Wang, Shaohe Lv, Xiaodong Wang, and Xinye Lin.
\newblock Gated convolutional {LSTM} for speech commands recognition.
\newblock In {\em {Proceedings of the International Conference of Computer
  Science}}, pages 669--681, June 2018.

\bibitem{Tara2015}
Tara~N. Sainath and Carolina Parada.
\newblock Convolutional neural networks for small-footprint keyword spotting.
\newblock In {\em {Proceedings of the annual conference of the International
  Speech Communication Association, INTERSPEECH}}, pages 1478--1482. {ISCA},
  September 2015.

\bibitem{Zhang2017}
Yundong Zhang, Naveen Suda, Liangzhen Lai, and Vikas Chandra.
\newblock Hello edge: Keyword spotting on microcontrollers.
\newblock {\em Computing Research Repository CoRR}, abs/1711.07128, November
  2017.

\bibitem{McMahan2018}
Brian McMahan and Delip Rao.
\newblock Listening to the world improves speech command recognition.
\newblock {\em Computing Research Repository CoRR}, abs/1710.08377, October
  2017.

\bibitem{FChollet2017}
F.~Chollet.
\newblock Xception: Deep learning with depthwise separable convolutions.
\newblock In {\em Proceedings of the IEEE Conference on Computer Vision and
  Pattern Recognition (CVPR)}, pages 1800--1807, July 2017.

\bibitem{Warden2018}
Pete Warden.
\newblock Speech commands: {A} dataset for limited-vocabulary speech
  recognition.
\newblock {\em Computing Research Repository CoRR}, abs/1804.03209, April 2018.

\bibitem{Szymon2016}
Szymon Bus and Konrad Jedrzejewski.
\newblock Digital signal processing techniques for pitch shifting and time
  scaling of audio signals.
\newblock In {\em Proceedings of SPIE - The International Society for Optical
  Engineering}, volume 10031, pages 1003157--1, September 2016.

\bibitem{Liu2019}
Chenxi Liu, Liang{-}Chieh Chen, Florian Schroff, Hartwig Adam, Wei Hua, Alan~L.
  Yuille, and Li~Fei{-}Fei.
\newblock Auto-deeplab: Hierarchical neural architecture search for semantic
  image segmentation.
\newblock {\em Computing Research Repository CoRR}, abs/1901.02985, January
  2019.

\bibitem{Song2018}
L.~{Song}, J.~{Liu}, B.~{Qian}, M.~{Sun}, K.~{Yang}, M.~{Sun}, and S.~{Abbas}.
\newblock A deep multi-modal {CNN} for multi-instance multi-label image
  classification.
\newblock {\em IEEE Transactions on Image Processing}, 27(12):6025--6038,
  December 2018.

\bibitem{Zheng2018}
Zheng Xu, Xitong Yang, Xue Li, and Xiaoshuai Sun.
\newblock The effectiveness of instance normalization: a strong baseline for
  single image dehazing.
\newblock {\em Computing Research Repository CoRR}, abs/1805.03305, May 2018.

\bibitem{Yunhui2019}
Yunhui {Guo}, Yandong {Li}, Rogerio {Feris}, Liqiang {Wang}, and Tajana
  {Rosing}.
\newblock {Depthwise Convolution is All You Need for Learning Multiple Visual
  Domains}.
\newblock {\em Association for the Advancement of Artificial Intelligence},
  February 2019.

\bibitem{Howard2017}
Andrew~G. Howard, Menglong Zhu, Bo~Chen, Dmitry Kalenichenko, Weijun Wang,
  Tobias Weyand, Marco Andreetto, and Hartwig Adam.
\newblock Mobilenets: Efficient convolutional neural networks for mobile vision
  applications.
\newblock {\em Computing Research Repository CoRR}, abs/1704.04861, April 2017.

\bibitem{Goodfellow2016}
Ian Goodfellow, Yoshua Bengio, and Aaron Courville.
\newblock {\em Deep Learning}.
\newblock MIT Press, 2016.
\newblock \url{http://www.deeplearningbook.org}.

\bibitem{Kingma14}
Diederik~P. Kingma and Jimmy Ba.
\newblock Adam: A method for stochastic optimization.
\newblock In {\em 3rd International Conference of Learning Representations
  (ICLR)}, December 2014.

\bibitem{Jaderberg2016}
Max Jaderberg, Volodymyr Mnih, Wojciech~Marian Czarnecki, Tom Schaul, Joel~Z
  Leibo, David Silver, and Koray Kavukcuoglu.
\newblock Reinforcement learning with unsupervised auxiliary tasks.
\newblock {\em arXiv preprint arXiv:1611.05397}, 2016.

\end{thebibliography}

\end{footnotesize}


\end{document}